\documentclass[final]{cvpr}

\usepackage{times}
\usepackage{epsfig}
\usepackage{graphicx}
\usepackage{amsmath}
\usepackage{amssymb}

\usepackage{comment}
\usepackage{color}

\usepackage{bm}
\usepackage{multirow}
\usepackage{booktabs}
\usepackage{graphicx}
\usepackage{wrapfig}
\usepackage{enumerate}
\usepackage{subfigure}
\usepackage{algorithm}
\usepackage{algpseudocode}
\usepackage{bbding}

\usepackage[pagebackref=true,breaklinks=true,colorlinks,bookmarks=false]{hyperref}
\usepackage[numbers, sort]{natbib}

\def\cvprPaperID{1198} % *** Enter the CVPR Paper ID here
\def\confYear{CVPR 2021}

\begin{document}

%%%%%%%%% TITLE
\title{Meta-Mining Discriminative Samples for Kinship Verification}

\author{Wanhua Li$^{1,2}$, Shiwei Wang$^{3}$, Jiwen Lu$^{1,2,\ast}$, Jianjiang Feng$^{1,2}$, Jie Zhou$^{1,2}$\\
$^{1}$Department of Automation, Tsinghua University, China \\
$^{2}$Beijing National Research Center for Information Science and Technology, China \\
$^{3}$School of Modern Post, Beijing University of Posts and Telecommunications\\
{\tt\small li-wh17@mails.tsinghua.edu.cn \qquad 2018110842@bupt.edu.cn}\\
{\tt \small   \{lujiwen,jfeng,jzhou\}@tsinghua.edu.cn}
}

\maketitle

\pagestyle{empty}  % no page number for the second and the later pages
\thispagestyle{empty} % no page number for the first page

\let\thefootnote\relax\footnotetext{$^{\ast}$ Corresponding author}

%%%%%%%%% ABSTRACT
\begin{abstract}
Kinship verification aims to find out whether there is a kin relation for a given pair of facial images. Kinship verification databases are born with unbalanced data. For a database with $N$ positive kinship pairs, we naturally obtain $N(N-1)$ negative pairs. How to fully utilize the limited positive pairs and mine discriminative information from sufficient negative samples for kinship verification remains an open issue. To address this problem, we propose a Discriminative Sample Meta-Mining (DSMM) approach in this paper. Unlike existing methods that usually construct a balanced dataset with fixed negative pairs, we propose to utilize all possible pairs and automatically learn discriminative information from data. Specifically, we sample an unbalanced train batch and a balanced meta-train batch for each iteration. Then we learn a meta-miner with the meta-gradient on the balanced meta-train batch. In the end, the samples in the unbalanced train batch are re-weighted by the learned meta-miner to optimize the kinship models. Experimental results on the widely used KinFaceW-I, KinFaceW-II, TSKinFace, and Cornell Kinship datasets demonstrate the effectiveness of the proposed approach.
\end{abstract}

%%%%%%%%% BODY TEXT
\section{Introduction}
Facial appearance conveys valuable information, such as identity~\cite{deng2019arcface,liu2017sphereface}, age~\cite{li2019bridgenet,wen2020adaptive}, gender~\cite{levi2015age,rodriguez2017age}, emotions~\cite{wang2020suppressing,kossaifi2020factorized}, social relation~\cite{zhang2015learningiccv,li2020grapheccv}, and so on. Recently, a variety of efforts~\cite{lu2013neighborhood,kohli2016hierarchical,wang2020kinship,robinson2018visual,liang2018weighted} have been devoted to kinship verification, which predicts the existence of kinship for a given pair of facial images. As an emerging task, it has many potential applications including missing children search~\cite{lu2013neighborhood}, intelligent family album organization~\cite{kohli2016hierarchical,qin2015tri}, and social media analysis~\cite{xu2014social,zhou2012gabor}.

\begin{figure}[t]
  \centering
  \subfigure[Existing Method]{
    \label{fig:motivation:a}
    \includegraphics[width=1.0\linewidth]{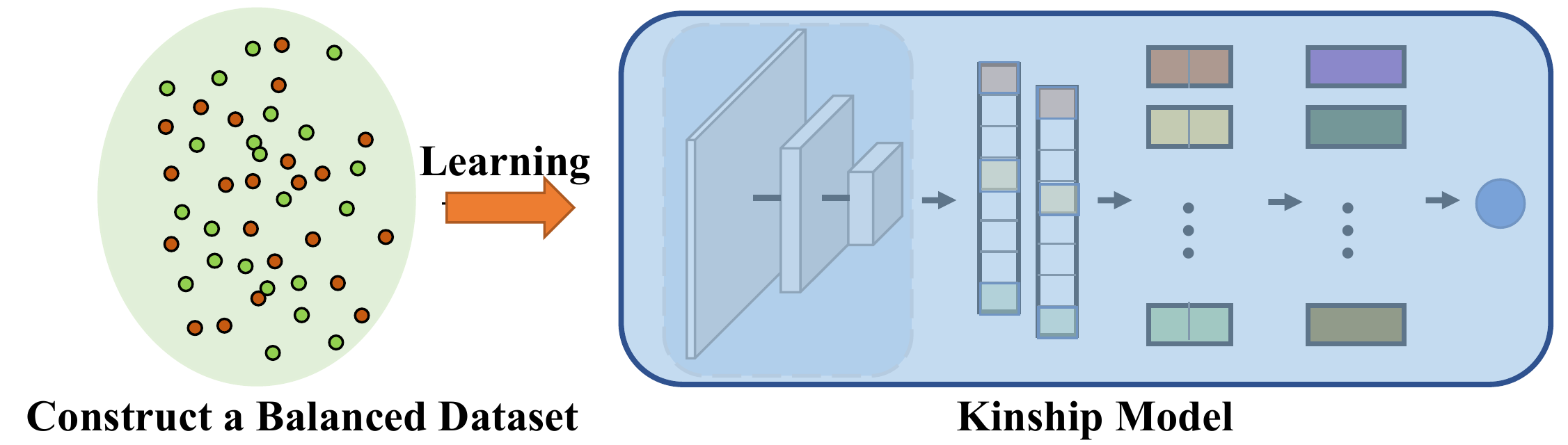}
  }
  \subfigure[Our Method]{
    \label{fig:motivation:b}
    \includegraphics[width=1.0\linewidth]{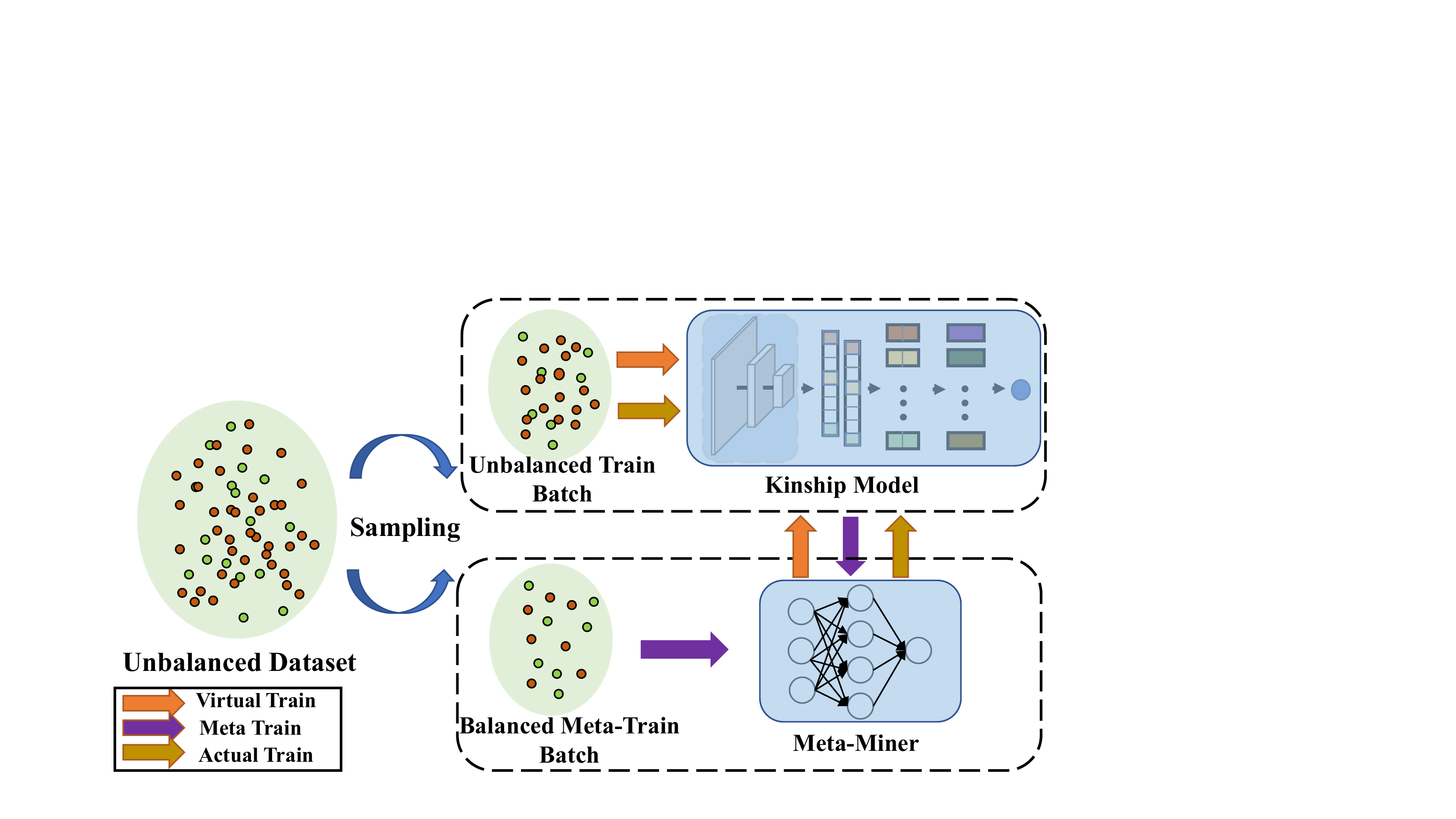}
  }
  \caption{The key idea of our method. (a) The existing method constructs a fixed database by sampling the equal size of negative samples. Then a kinship model is learned with the balanced database. (b) Our method utilizes all possible pairs and does not discard any negative samples.  Specifically, we sample an unbalanced train batch and a balanced meta-train batch for each iteration. A meta-miner is introduced to mine the samples in the training batch. Our method alternately optimizes the kinship model and the proposed meta-miner via a meta-learning framework .}
  \label{fig:motivation}
\vspace{-0.5cm}
\end{figure}

Existing kinship databases organize data in terms of positive samples. For example, KinFaceW-I and KinFaceW-II~\cite{lu2013neighborhood} datasets collect parent-child image pairs from the Internet search. Assuming that a kinship database contains $N$ positive kinship pairs, the negative samples are generated by combining all unrelated parent-child image pairs. Therefore, we obtain $N(N-1)$ negative samples, which is far more than the number of positive pairs. Existing methods usually randomly sample fixed $N$ negative samples to construct a balanced database. However, this strategy simply ignores the remaining $N(N-2)$ negative samples leading to overfitting. Besides, the real decision boundary cannot be well learned with randomly selected $N$ negative samples. One simple strategy to address this issue is to sample a balanced batch from positive samples and all possible negative pairs separately. Nevertheless, most negative pairs are easy samples and they contribute little to the network training.

In this work, we investigate how to mine discriminative information from limited positive pairs and sufficient negative samples and propose a Discriminative Sample Meta-Mining (DSMM) strategy via a meta-learning framework. Specifically  We first randomly sample an unbalanced train batch from all possible pairs with a positive to negative ratio of $1:C$, where $C > 1$. Then we aim to mine the discriminative samples by re-weighting these samples in the training batch. Instead of manually selecting fixed weighting functions, we consider automatically learning the weighting functions from data. Encouraged by the recent success of meta-learning~\cite{finn2017model,vinyals2016matching,sung2018learning}, we introduce a meta-miner network, which predicts the weight for each sample and is optimized with the meta-gradient. Concretely, another balanced meta-train batch is sampled from all possible pairs. Then we optimize the kinship model and meta-miner network alternately with these two batches. After one-step-forward optimization for the kinship model on the unbalanced train batch, we train the meta-miner network on the balanced meta-train batch with one-step-forward parameters of the kinship model. In the end, we perform the real optimization for the kinship model with the sample weights generated by the updated meta-miner network. Figure \ref{fig:motivation} shows the key idea of our method. The contributions of this paper are summarized as follows:
\begin{itemize}
\item
  To the best of our knowledge, the proposed discriminative sample meta-mining approach is the first attempt to fully utilize the unbalanced data of kinship databases via a meta-learning framework.
\item
  Our method proposes to simultaneously sample an unbalanced train batch and a balanced meta-train batch, and then perform sample mining on the training batch with the guidance of a balanced meta-train batch.
\item
  Extensive experiments on four widely used kinship databases illustrate that the proposed method achieves state-of-the-art results.
\end{itemize}

\section{Related Work}

\textbf{Kinship Verification:} Over the past decade, a variety of approaches~\cite{fang2010towards,xia2012understanding,lu2014neighborhood,liang2018weighted} have been proposed for facial kinship verification. We can categorize them into three groups~\cite{li2020graph}: hand-crafted methods, metric-learning based methods, and deep learning-based methods. In traditional exploration, hand-crafted methods have been widely used, such as histogram of the gradient~\cite{fang2010towards,somanath2012can,zhou2011kinship}, Gabor gradient orientation pyramid~\cite{zhou2012gabor},  self-similarity~\cite{kohli2012self}, and so on.
%Fang \emph{et al.}~\cite{fang2010towards} proposed a  pictorial structure model with 22 extracted features including facial part colors, facial distances, and histogram of gradients feature.  Kohli \emph{et al.}~\cite{kohli2012self} normalized the detected faces using Weber's law based normalization technique to remove the illumination factor.
However, hand-crafted methods usually suffer from the limited performance. Distance metric based approach has become the most popular and successful method due to the superior performance in kinship verification~\cite{lu2014neighborhood,lu2017discriminative}. For example,
%Lu \emph{et al.}\cite{lu2017discriminative} proposed a discriminative deep metric learning method, which learned a set of hierarchical nonlinear transformations to address the nonlinear and scalability issues.
Liang \emph{et al.}~\cite{liang2018weighted} presented a weighted graph embedding-based metric learning method, which jointly learned multiple metrics by constructing an intrinsic graph. Recent years have witnessed the remarkable success of deep learning in computer vision, such as 
%image recognition~\cite{tan2019mnasnet,tan2019efficientnet},
object detection~\cite{carion2020end,peng2020ida}, and face recognition~\cite{cao2020domain,huang2020curricularface}. Some researchers~\cite{zhang122015kinship,li2020graph} applied deep learning technologies into kinship verification. For example, Zhang \emph{et al.}~\cite{zhang122015kinship} extracted facial key-points features with CNNs.
%Li \emph{et al.}~\cite{li2020graph} further presented a graph-based method with graph neural networks.
However, all these methods only use a fixed database, where only a small proportion of the negative samples are used. By contrast, we aim to mine the discriminative samples from all possible pairs.

\textbf{Class Imbalance:} Most efforts to address the class imbalance issue can be grouped into two categories~\cite{zhou2020bbn}: re-sampling and re-weighting. The re-sampling strategy usually constructs a balanced batch by adding samples from minor classes~\cite{shen2016relay} or under-sampling the dominant classes~\cite{he2009learning}. The re-weighting strategy assigns different weights for samples from different classes in the loss function~\cite{ting2000comparative}. Many advanced class imbalance learning methods~\cite{lin2017focal,cui2019class} have been proposed in recent years. Lin \emph{et al.}~\cite{lin2017focal} proposed focal loss to deal with the extreme imbalance between foreground and background classes. The effective number of samples was introduced in~\cite{cui2019class} and a class-balanced loss was further proposed with a re-weighting scheme. Zhou \emph{et al.}~\cite{zhou2020bbn} proposed a Bilateral-Branch Network for long-tailed visual recognition, which considers representation learning and classifier learning simultaneously.

\textbf{Meta-Learning:} Meta-learning aims to develop a machine learning system that is used to assist in the optimization of other machine learning models~\cite{finn2017model,ren2018learning,shu2019meta,vinyals2016matching}. Some researchers have proposed optimization based meta-learning approaches~\cite{andrychowicz2016learning,chen2017learning,ravi2016optimization,finn2017model}. Andrychowicz \emph{et al.}~\cite{andrychowicz2016learning} developed an algorithm with LSTMs to replace the hand-designed optimization algorithms.
%A model-agnostic meta-learning algorithm was proposed in~\cite{finn2017model}, which learned a good initialization to enable the quick adaptation.
A variety of metric-based methods~\cite{vinyals2016matching,snell2017prototypical} have been proposed, especially for the area of one/few shot learning.
%Snell \emph{et al.}~\cite{snell2017prototypical} presented the prototypical networks for the problem of few-shot classification, where the prototype representations of each class are computed in the learned metric space.
Sung \emph{et al.}~\cite{sung2018learning} utilized a relation network to learn an embedding and a deep distance metric for comparing two samples. There are also studies~\cite{santoro2016meta,mishra2018simple} focusing on memory-based approaches. Santoro \emph{et al.}~\cite{santoro2016meta} proposed a memory-augmented neural network to rapidly encode new data and make accurate predictions with the assimilated data. Encouraged by the success of meta-learning, we design a meta-miner and utilize it to guide the training process of our kinship model by fully exploiting all possible samples in the database. Although meta-learning has been successfully applied to face recognition~\cite{guo2020learning}, our method is still significantly different from it in terms of motivation, framework, optimization strategy, \emph{etc.}

\section{Proposed Approach}
In this section, we first introduce the unbalanced data issue for kinship verification. Then, we detail how the proposed discriminative sample meta-mining approach exploits limited positive samples and sufficient negative samples. The meta-learning framework is further presented. Lastly, we introduce the network design of our kinship relation model and the proposed meta-miner.

\renewcommand{\algorithmicrequire}{\textbf{Input:}} % Use Input in the format of Algorithm
\renewcommand{\algorithmicensure}{\textbf{Output:}} % Use Output in the format of Algorithm
\begin{algorithm}[t]
\caption{\!\!\! \textbf{:} Discriminative Sample Meta-Mining.}
\label{alg:DSMM}
\begin{algorithmic}[1]
\Require
Iteration number $T$.
\Ensure
Model parameters $\theta^T$ and $\phi^T$.
\State Initialize model parameters $\theta^0$ and $\phi^0$.
\For{$t = 0,1,...,T-1$ }
\State Sample an unbalanced train batch $ (\mathcal{S}^{trn,p},\mathcal{S}^{trn,n})$.
\State Sample a balanced batch $ (\mathcal{S}^{meta,p},\mathcal{S}^{meta,n})$.
\State Perform the virtual training by \eqref{equ:loss:total2} - \eqref{equ:update:virtual}
\State Update $\phi^{t+1}$ on the meta-train batch via \eqref{equ:loss:meta} - \eqref{equ:update:meta}.
\State Normalize sample weights with \eqref{equ:normalize}.
\State Update $\theta^{t+1}$ by \eqref{equ:loss:actual} - \eqref{equ:update:actual}.
\EndFor
\\
\Return  $\theta^T$, $\phi^T$.
\end{algorithmic}
\end{algorithm}

\subsection{Unbalanced Kinship Data}
Kinship verification organizes each sample as a facial image pair $(\bm{x},\bm{y})$, where $\bm{x}$ denotes the parent image and $\bm{y}$ is the child image. Existing kinship verification databases are usually constructed by collecting positive kinship pairs since the negative pairs can be generated by shuffling the positive pairs. Assuming that there are a total of $N$ positive pairs in a dataset, then we naturally obtain $N(N-1)$ negative samples. Therefore, the size of the negative samples is much larger than that of positive samples.

How to learn a kinship model with the unbalanced data remains an open issue. Most existing methods randomly select $N$ negative samples to form a balanced dataset and discard the remaining $N(N-2)$ negative samples. Formally, we denote the constructed positive data and negative data by $\mathcal{D}^{trn,p}$ and  $\mathcal{D}^{trn,n}$, respectively. Let $f(\bm{x},\bm{y};\theta)$ denote the kinship model which is parameterized by $\theta$. Then the model is trained with binary cross entropy loss on the balanced dataset:
\begin{equation}
\begin{aligned}
\mathcal{L} =   &-\frac{1}{2N}  \sum_{(\bm{x},\bm{y}) \in \mathcal{D}^{trn,p}} \log(f(\bm{x},\bm{y};\theta)) \\
&-\frac{1}{2N} \sum_{(\bm{x},\bm{y}) \in \mathcal{D}^{trn,n}} \log(1 - f(\bm{x},\bm{y};\theta)).
\end{aligned}
\label{equ:loss:common}
\end{equation}

However, these methods cannot fully utilize the negative data leading to inaccurate decision boundaries. One strategy to address this issue is to sample a balanced mini-batch for each iteration from all possible pairs.  Formally, a positive batch $\mathcal{B}^{trn,p}$ with $m$ examples and a negative batch $\mathcal{B}^{trn,n}$ with $m$ pairs are sampled. Then we train the kinship model as follows:
\begin{equation}
\begin{aligned}
\mathcal{L} =   &-\frac{1}{2m}  \sum_{(\bm{x},\bm{y}) \in \mathcal{B}^{trn,p}} \log(f(\bm{x},\bm{y};\theta)) \\
&-\frac{1}{2m} \sum_{(\bm{x},\bm{y}) \in \mathcal{B}^{trn,n}} \log(1 - f(\bm{x},\bm{y};\theta)).
\end{aligned}
\label{equ:loss:balance}
\end{equation}
This strategy can use all possible pairs with a balanced batch. Nevertheless, most negative pairs are easy examples, which contribute little to the network training. In this paper, we aim to mine the discriminative information from the limited positive samples and sufficient negative facial image pairs simultaneously.

\subsection{Discriminative Sample Meta-Mining}

To address the above issue, we propose a discriminative sample meta-mining approach with a meta-learning framework. To fully exploit the potential of kinship data, we perform sampling from all possible pairs to construct a batch. While the kinship data is unbalanced and most negative samples are easy ones, we manually construct an unbalanced train batch with a positive to negative ratio of $1:C (C > 1)$. We then mine the discriminative samples from the unbalanced batch. We employ a sample re-weighting strategy to mining the discriminative information, which re-weights the loss value for each sample. One can assign a constant value or choose a hand-designed weighting function to re-weight the samples in the unbalanced train batch. Instead, we aim to learn a weighting function directly from data with a meta-learning framework. Specifically, we introduce a meta-miner to generate the weights. We implement the meta-miner with a neural network which is denoted as $g(\phi)$ parameterized by $\phi$. The positive set and negative set of the sampled train batch are denoted as $\mathcal{S}^{trn,p}$ and  $\mathcal{S}^{trn,n}$, respectively. Then the training objective for a given fixed meta-miner network $g(\phi)$ is defined as:
\begin{equation}
\begin{aligned}
\mathcal{L} =   &\frac{-1}{m(1+C)} ( \sum_{\bm{s} \in \mathcal{S}^{trn,p}} g(\bm{s};\phi) \log(f(\bm{s};\theta)) \\
& + \sum_{\bm{s} \in \mathcal{S}^{trn,n}} g(\bm{s};\phi) \log(1 - f(\bm{s};\theta))),
\end{aligned}
\label{equ:loss:total}
\end{equation}
where $m$ is the size of positive samples in the training batch and $\bm{s}$ represents a pair of positive images.

We then optimize the parameters $\theta$ and $\phi$ with an online strategy. The framework is depicted in Figure \ref{fig:motivation:b}.

\begin{figure*}[t]
  \centering
  \includegraphics[width=1.0\linewidth]{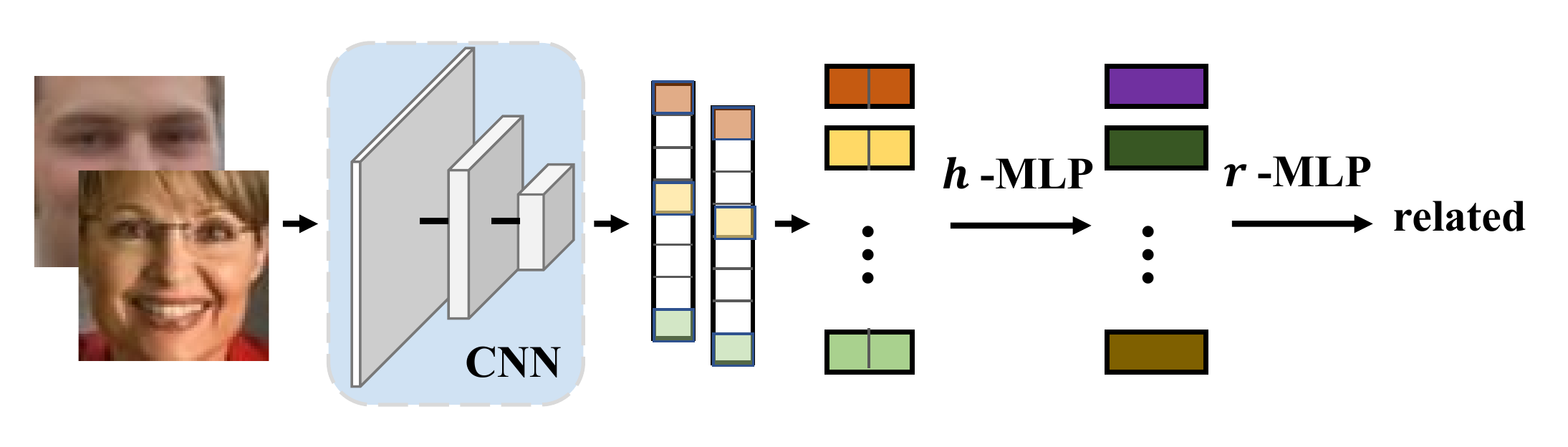}
  \caption{The overall framework of our kinship relation model. For paired facial images, we use a CNN to extract deep features for them. Then we explicitly model the relations of these two features with a relation module.}
  \label{fig:kinship}
  \vspace{-0.5cm}
\end{figure*}

\textbf{Virtual Training the Kinship Model:} Assuming that we have obtained the parameters $\theta^t$ and $\phi^t$ at $t$-th time step, now we perform a virtual update for $\theta^t$ to obtain $\hat{\theta}^{t+1}$. The unbalanced train batch is used for updating. Therefore, the loss function for this virtual training step is expressed as:
\begin{equation}
\begin{aligned}
\mathcal{L}_{v\_trn}(\phi^t) =   &\frac{-1}{m(1+C)} ( \sum_{\bm{s} \in \mathcal{S}^{trn,p}} g(\bm{s};\phi^t) \log(f(\bm{s};\theta^t)) \\
& + \sum_{\bm{s} \in \mathcal{S}^{trn,n}} g(\bm{s};\phi^t) \log(1 - f(\bm{s};\theta^t))).
\end{aligned}
\label{equ:loss:total2}
\end{equation}
Since different $\phi^t$ leads to different virtual train loss values, the loss in \eqref{equ:loss:total2} is essentially a function of  $\phi^t$. With the defined loss, we can update the kinship model with SGD:
\begin{equation}
\begin{aligned}
\hat{\theta}^{t+1}(\phi^t) = \theta^t - \alpha \nabla_{\theta^t} \mathcal{L}_{v\_trn}(\phi^t),
\end{aligned}
\label{equ:update:virtual}
\end{equation}
where $\alpha$ is the step size. It is worth noting that the virtual training step does not actually update the parameter $\theta^t$ of the kinship model, which is used to calculate the second-order gradients in the next step.

\textbf{Training the Meta-Miner Network:} Having obtained the $\hat{\theta}^t$, we then train the meta-miner network to obtain $\phi^{t+1}$. Specifically, we optimize the parameter of the meta-miner network using the meta-learning idea~\cite{finn2017model,andrychowicz2016learning}. Another balanced meta train batch is sampled, which is used to guide the training process of the meta-miner network. The motivation of this step is that we expect that the obtained parameter $\hat{\theta}^{t+1}$ in the last step can achieve better performance on the balanced meta-train batch. Formally, we define the positive set and the negative set of the balanced meta-train batch as $\mathcal{S}^{meta,p}$ and $\mathcal{S}^{meta,n}$, respectively. Then the meta-loss is given by:
\begin{equation}
\begin{aligned}
\mathcal{L}_{meta}(\phi^t) =   &-\frac{1}{2m}  \sum_{\bm{s} \in \mathcal{S}^{meta,p}}  \log(f(\bm{s};\hat{\theta}^{t+1}(\phi^t))) \\
&-\frac{1}{2m} \sum_{\bm{s} \in \mathcal{S}^{meta,n}} \log(1 - f(\bm{s};\hat{\theta}^{t+1}(\phi^t)))),
\end{aligned}
\label{equ:loss:meta}
\end{equation}
where we use $m$ positive samples and $m$ negative samples to form a balanced batch. Then we update the parameter $\phi^t$ via gradient descent method with a learning rate of $\beta$:
\begin{equation}
\begin{aligned}
\phi^{t+1} = \phi^t - \beta \nabla_{\phi^t} \mathcal{L}_{meta}(\phi^t).
\end{aligned}
\label{equ:update:meta}
\end{equation}

With the updated the parameter $\phi^{t+1}$, we can generate the real weights for the samples in the unbalanced train batch. For each sample $\bm{s} \in \mathcal{S}^{trn}$, we obtain the weight $g(\bm{s},\phi^{t+1})$. Before they are used in the next step, we first normalize them within the train batch:
\begin{equation}
\begin{aligned}
\widetilde{w}_{\bm{s}}= \frac{g(\bm{s};\phi^{t+1})}{\sum_{\bm{q} \in \mathcal{S}^{trn,p} } g(\bm{q};\phi^{t+1}) + \sum_{\bm{q} \in \mathcal{S}^{trn,n} } g(\bm{q};\phi^{t+1})}.
\end{aligned}
\label{equ:normalize}
\end{equation}

\textbf{Actual Training the Kinship Model:} With the obtained weights in the second step, we can define the weighed loss function as follows:
\begin{equation}
\begin{aligned}
\mathcal{L}_{trn}(\theta^t) =   &\frac{-1}{m(1+C)} ( \sum_{\bm{s} \in \mathcal{S}^{trn,p}} \widetilde{w}_{\bm{s}} \log(f(\bm{s};\theta^t)) \\
& + \sum_{\bm{s} \in \mathcal{S}^{trn,n}} \widetilde{w}_{\bm{s}} \log(1 - f(\bm{s};\theta^t))).
\end{aligned}
\label{equ:loss:actual}
\end{equation}

Then we perform the actual network optimization for kinship model with gradient descent:
\begin{equation}
\begin{aligned}
\theta^{t+1} = \theta^t - \gamma \nabla_{\theta^t} \mathcal{L}_{trn}(\theta^t),
\end{aligned}
\label{equ:update:actual}
\end{equation}
where  $\gamma$ is the learning rate.

In this way, the parameters $\theta$ and $\phi$ are alternately optimized via meta-learning. We show the above framework in Algorithm \ref{alg:DSMM}.

\begin{figure*}[t]
  \centering
  \includegraphics[width=1.0\linewidth]{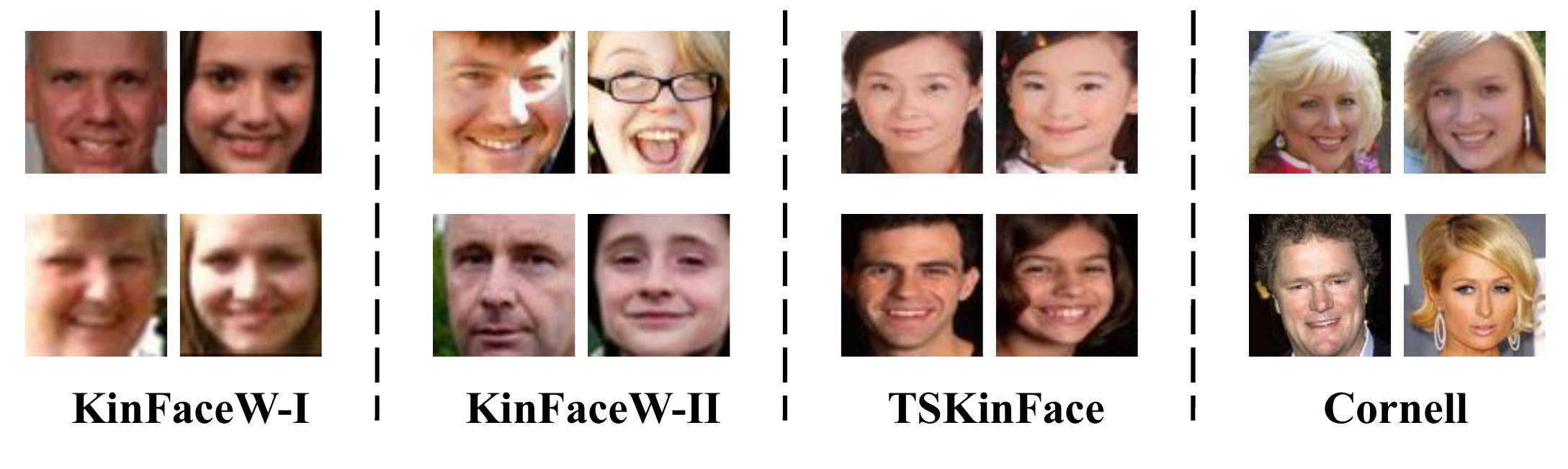}
  \caption{Some examples of four widely used databases. Each row shows a paired sample, where the first image represents the parent and the second image belongs to the child. }
  \label{dataset}
  \vspace{-0.5cm}
\end{figure*}

\subsection{Implementation Details}

The proposed DSMM method involves two networks: kinship model $f(\theta)$ and meta-miner network $g(\phi)$. Here we detail the design of these two models.

For the kinship model, we first use a ResNet-18 to extract image features for each facial image. Formally, for a paired sample $(\bm{x},\bm{y})$, we obtain the deep embeddings $(\bm{e}_{\bm{x}},\bm{e}_{\bm{y}}) \in (\mathbb{R}^D,\mathbb{R}^D)$, where $D$ represents the feature dimension. Inspired by the recent success of relation networks~\cite{santoro2017simple}, we consider modeling the relations of these two features to infer their kinship. We explicitly model the per-dimension relations of two extracted features and formulate the kinship relation model as follows:
\begin{equation}
\begin{aligned}
f(\theta) = r(\mathop{\big| \big|}_{i=1}^D h(\bm{e}_{\bm{x}}^i,\bm{e}_{\bm{y}}^i) ),
\end{aligned}
\label{equ:update:actual}
\end{equation}
where $\big| \big|$ represents concatenation, $\bm{e}_{\bm{x}}^i$ and $\bm{e}_{\bm{y}}^i$ denote the $i$-th element in the embeddings $\bm{e}_{\bm{x}}$ and $\bm{e}_{\bm{y}}$ respectively. Both $r(\cdot)$ and $h(\cdot)$ are implemented with MLPs. We depict the kinship relation model in Figure \ref{fig:kinship}.

For the meta-miner network $g(\phi)$, we implement it with a three-layer MLP. Instead of taking the whole samples as the input, we only use the loss information to generate the sample weights. Specifically, the meta-miner network takes sample labels, predictions of our kinship model, and corresponding train losses as inputs. Therefore, the input dimension of $g(\phi)$ is 3. Besides, we set the neural number of the hidden layer as 512, which achieves the best performance.

\begin{figure*}[t]
  \centering
  \subfigure[KinFaceW-I]{
    \label{fig:corre:morph2}
    \includegraphics[width=0.45\linewidth]{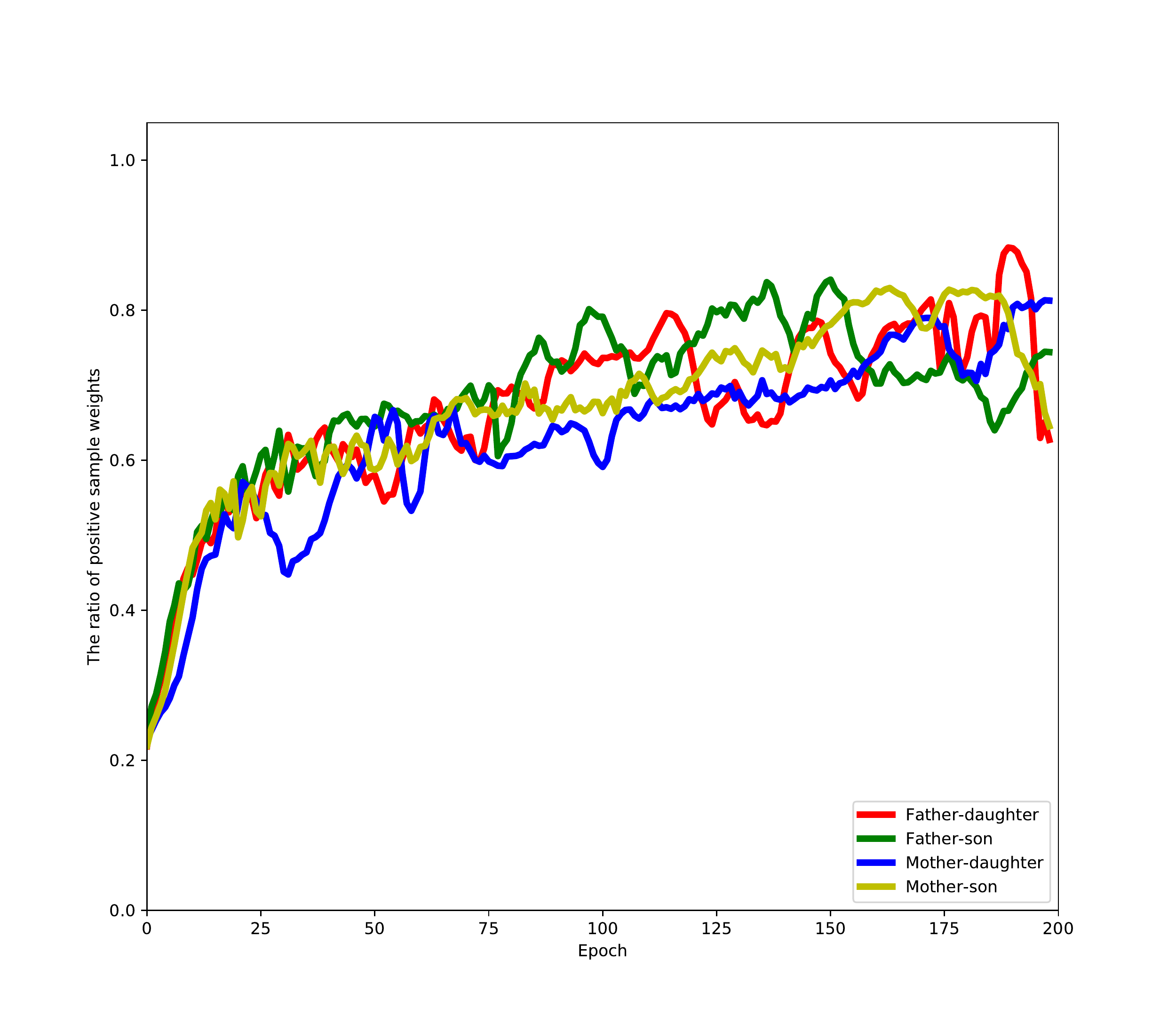}
  }
  \subfigure[KinFaceW-II]{
    \label{fig:corre:adience1}
    \includegraphics[width=0.45\linewidth]{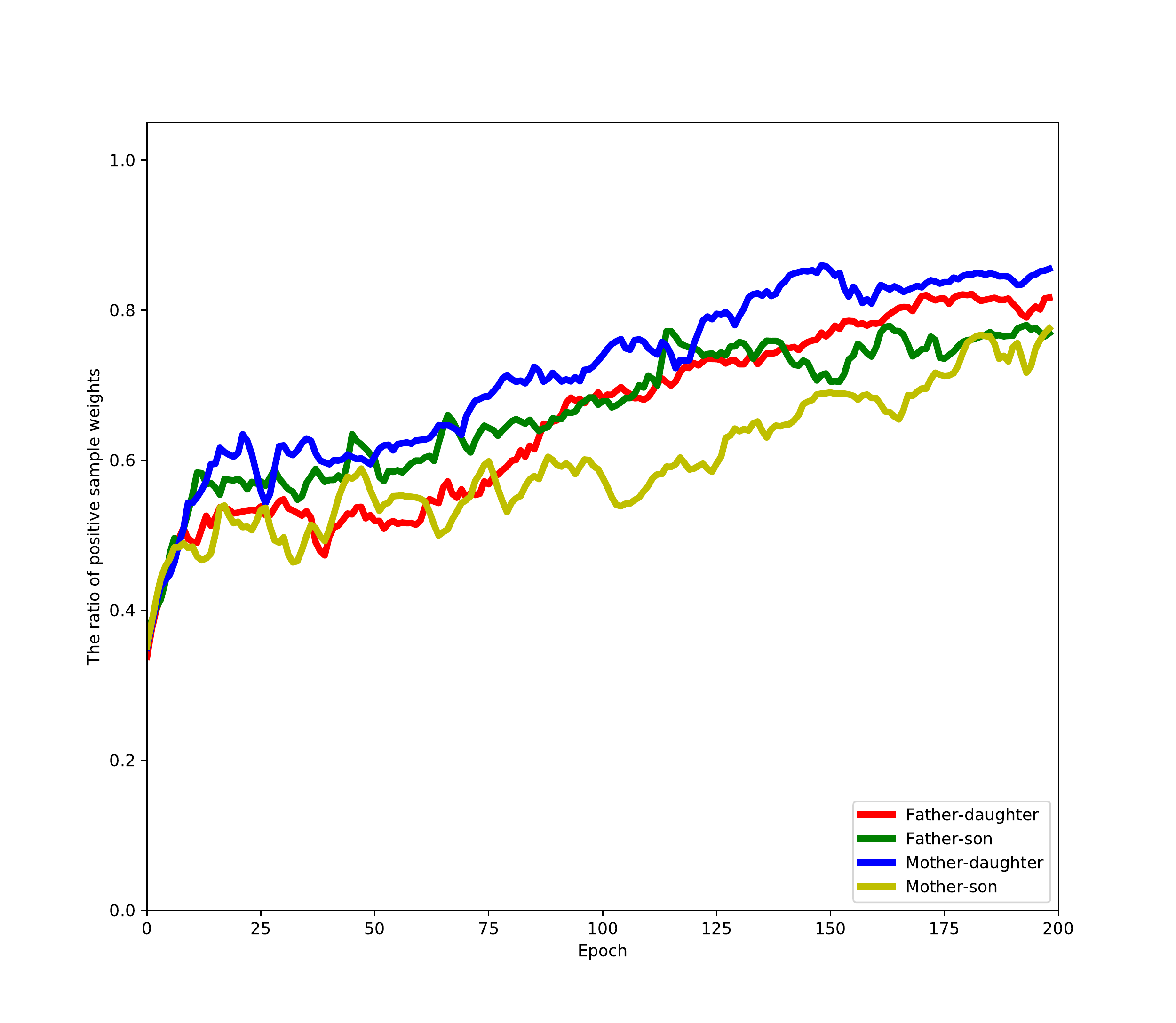}
  }
  \caption{The ratio curves of positive sample weights on the unbalanced train batch changing with epochs on the KinFaceW-I database and KinFaceW-II database.}
  \label{fig:ex1}
\vspace{-0.4cm}
\end{figure*}

\section{Experiments}
To evaluate the effectiveness of the proposed discriminative sample meta-mining approach, we conducted extensive experiments on four widely used kinship datasets
%, including the KinFaceW-I database~\cite{lu2014neighborhood}, KinFaceW-II database~\cite{lu2014neighborhood}, TSKinFace database~\cite{qin2015tri}, and Cornell~\cite{fang2010towards} KinFace database.

\subsection{Datasets}
{\bf KinFaceW-I Dataset~\cite{lu2014neighborhood}:} This dataset contains  533 pairs of facial images of persons with a kin relation. Four different kin relations are considered in the dataset: father and daughter (F-D) with 134 pairs, father and son (F-S) with 156 pairs, mother and daughter (M-D) with 127 pairs, mother and son (M-S) with 116 pairs. Each sample is composed of one parent face image and one child face image. The images are marked with the positions of eyes and are aligned to a size of $64\times64$ based on the landmarks. We follow the standard protocol~\cite{lu2014neighborhood} in this database which adopts the five-fold cross-validation.

{\bf KinFaceW-II Dataset~\cite{lu2014neighborhood}:} This dataset consists of 1000 pairs of facial images of individuals with a kin relation. This database also considers four common kin relations: father and daughter (F-D), father and son (F-S), mother and daughter (M-D), mother and son (M-S).  Different from the KinFaceW-I database, the positive pairs in this dataset are taken from the same photo. Following the standard protocol~\cite{lu2014neighborhood}, we perform five-fold cross-validation.

{\bf TSKinFace Dataset~\cite{wu2016usefulness}:} TSKinFace dataset has 2589 face images in total. Unlike the traditional kinship database which investigates bi-subject relations, it is constructed to model tri-subject relations. Two kin relations are considered in this database: Father-Mother-Son (FM-S) with 285 groups and Father-Mother-Daughter (FM-D) with 274 groups. Following~\cite{lu2017discriminative}, we re-organize the data into the four kinds of bi-subject relations as other datasets. Then we obtain  502 pairs of father and daughter (F-D) relations,  513 pairs of father and son (F-S) relations, 502 pairs of mother and daughter (M-D) relations, 513 pairs of mother and son (M-S) relations.  Five-fold cross-validation is adopted.

{\bf Cornell KinFace Dataset~\cite{fang2010towards}:} Cornell KinFace database is a small kinship dataset with only 150 pairs of face images. All images are collected from the internet including four family relations. There are 22\% father and daughter (F-D) relations, 40\% father and son (F-S) relations, 25\% mother and daughter (M-D) relations, and 13\% mother and son (M-S) relations in this dataset. We follow the standard protocol~\cite{fang2010towards} with five-fold cross-validation.

Some examples of these datasets are shown in Figure \ref{dataset}.

\begin{table}[t]
\caption{Mean accuracy (\%) of different fusion methods. EF means early fusion, LF represents late fusion.}
\label{table:fusion}
\vspace{0.2cm}
\renewcommand\tabcolsep{4pt}
\centering
\begin{tabular}{l|c|c|c|c|c}
\hline
Dataset  &  CNNP& EF & LF & Relation & DSMM  \\
\hline
KinFaceW-I & 77.5 &  75.7 & 72.3 & 78.1 & \textbf{82.4}   \\
\hline
KinFaceW-II & 88.4 & 87.9 & 81.8 & 90.4 & \textbf{93.0} \\
\hline
\end{tabular}
\vspace{-0.5cm}
\end{table}

\subsection{Experimental Settings}
For all datasets, we performed data augmentation to alleviate the overfitting issue. First, we resized the image size to $73\times73$, and then randomly cropped it to a size of $64\times64$. Finally, we performed a random horizontal flipping with a probability of 50\%. The positive sample batch size on the KinFaceW-II and TSKinFace datasets was set 16. The positive sample batch size on the KinFaceW-I and Cornell KinFace datasets was set to 8 since they have smaller data sizes. For the kinship relation model, we used a ResNet-18 network to extract image features, which was trained with binary cross-entropy loss. We optimized our network with Adam optimizer.  We set the learning rates $\alpha=0.001$, $\beta=0.0001$, and $\gamma=0.001$. We decayed the learning rate $\gamma$ by 0.1 at 100 and 150 epochs. We trained our model for 200 epochs. The meta-miner network was implemented with an MLP. The output dimension of $g(\phi)$ was set to 1, which corresponds to the generated sample weight. A sigmoid function was employed to normalize the output.

\subsection{Results and Analysis}
%In this subsection, we present the experimental results and detailed analisys on four widely used kinship databases.
In this subsection, we present the experimental results on four kinship databases and give a detailed analysis.

\begin{table}[t]
\centering
\caption{Ablation study on the KinFaceW-I dataset.}
\label{tab:abaKFW1}
\vspace{0.2cm}
\begin{tabular}{cccccc}
\toprule
Method  & FD & FS & MD & MS & Mean\\
\midrule
Balance Batch & 76.2 & 79.5 & 87.9 & 79.8 & 80.9 \\
Unbalance Batch & 75.8 & 80.1 & 88.2 & 79.3 & 80.9 \\
Balance + Focal & 69.8 & 72.1 & 83.5 & 72.8 & 74.6 \\
Unbalance + Focal & 71.3 & 73.7 & 83.0 & 75.4 & 75.8 \\
\midrule
DSMM& {\bf76.7} & {\bf81.7} & {\bf89.0} & {\bf82.3} & {\bf82.4} \\
\bottomrule
\end{tabular}
%\vspace{-0.2cm}
\end{table}

\begin{table}[t]
\centering
\caption{Ablation study on the KinFaceW-II dataset.}
\label{tab:abaKFW2}
\vspace{0.2cm}
\begin{tabular}{cccccc}
\toprule
Method  & FD & FS & MD & MS & Mean\\
\midrule
Balance Batch & 89.2 & 92.0 & 94.2 & 92.0 & 91.8 \\
Unbalance Batch  & 89.2 & 92.0 & 94.8 & {\bf94.0} & 92.5 \\
Balance + Focal & 78.6 & 79.0 & 83.0 & 81.4 & 80.5 \\
Unbalance + Focal & 79.4 & 81.6 & 86.4 & 84.4 & 83.0 \\
\midrule
DSMM& {\bf89.8} & {\bf92.6} & {\bf95.8} & 93.6 & {\bf93.0} \\
\bottomrule
\end{tabular}
\vspace{-0.5cm}
\end{table}

{\bf Kinship Relation Model:} Note that the proposed kinship relation model as defined in \eqref{equ:update:actual} is novel to kinship verification. Since kinship verification deals with two facial images, one key design is how to fuse these two images. The closest method is CNNP~\cite{zhang122015kinship}, which uses a self-designed backbone and early fusion. We consider several different strategies and show the results in Table \ref{table:fusion}. Compared with CNNP, the EF (early fusion) solution only replaces the feature extractor model with ResNet-18, which achieves slightly worse performance due to a simper network. For simplicity, the ResNet-18 is adopted as the backbone for all other strategies. The LF (late fusion) solution fuses two images in the last convolution layer instead of the input layer, leading to terrible results. As a late fusion strategy, our proposed relation module greatly improves performance, which shows the effectiveness of our kinship relation model. With the proposed kinship relation model, our DSMM further improves the accuracy by a large margin.

{\bf Ablation Study:}
To validate the effectiveness of our method, we consider several different strategies to handle the unbalanced data issue. The first strategy is named Balance Batch, which is a re-sampling strategy. It randomly selects a balanced batch from all possible pairs and uses the sampled balance batch to train our kinship relation model. The second method sample the same unbalanced train batch as our method does. To address the unbalanced issue, we directly set the sample weight for all negative samples as $1/C$, which is a re-weighting strategy. We also consider applying focal loss~\cite{lin2017focal} for the sampled balanced batches or unbalanced batches to mine discriminative samples.

We conducted experiments with these different strategies on the KinFaceW-I and KinFaceW-II datasets. The results are shown in Table~\ref{tab:abaKFW1} and Table~\ref{tab:abaKFW2}, respectively. We observe that Unbalance Batch usually gives better performance. The reason is that an unbalanced batch contains more negative samples so that they can be mined later. Since negative samples usually are the easy ones, sampling more negative samples are more useful. Both Balance Batch and Unbalance Batch can improve performance compared with the kinship relation model baseline as shown in Table~\ref{table:fusion}. However, the Focal loss significantly hurts performance. Due to the limited data size, we observed significant over-fitting (the accuracy on the training set is close to 100\%). The use of Focus Loss makes only a few hard samples dominate the gradient, resulting in a highly biased model with poor generalization. By contrast, our method introduces the balanced meta-train batch to guarantee the generalization ability of our model. Besides, our method can dynamically fit a wide range of weighting functions rather than just a certain family of functions. In the end, we see that our method achieves the best performance on both datasets, which demonstrates the effectiveness of our proposed approach.

\begin{table}[t]
\centering
\caption{Results with different $C$ on the KinFaceW-I dataset.}
\label{tab:KFW1C}
\vspace{0.2cm}
\renewcommand\tabcolsep{10pt}
\begin{tabular}{cccccc}
\toprule
$C$  & FD & FS & MD & MS & Mean\\
\midrule
1& 76.6 & 80.1 & 88.6 & 81.9 & 81.8 \\
2 & 75.3 & 81.1 & 86.9 & 78.9 & 80.6\\
4& {\bf76.7} & {\bf81.7} & {\bf89.0} & {\bf82.3} & {\bf82.4} \\
8 & 75.5 & 77.9 & 87.8 & 78.0 & 79.8\\
\bottomrule
\end{tabular}
%\vspace{-0.2cm}
\end{table}

\begin{table}[t]
\centering
\renewcommand\tabcolsep{10pt}
\caption{Results with different $C$ on the  KinFaceW-II dataset.}
\label{tab:KFW2C}
\vspace{0.2cm}
\begin{tabular}{cccccc}
\toprule
$C$  & FD & FS & MD & MS & Mean\\
\midrule
1 & 89.0 & 91.8 & {\bf96.0} & 93.4 & 92.5 \\
2& {\bf89.8} & {\bf92.6} & 95.8 & {\bf93.6} & {\bf93.0} \\
3 & 87.6 & 91.4 & 94.0 & 92.0 & 91.3\\
4 & 87.0 & 91.0 & 92.8 & 91.2 & 90.5\\
\bottomrule
\end{tabular}
\vspace{-0.5cm}
\end{table}

To better understand the learning process of our meta-miner, we show the ratio of positive sample weights on the unbalanced train batch, which is defined as the total of positive weights divided by the total of all sample weights per batch. The curves are depicted in Figure ~\ref{fig:ex1}. We observe that the ratio of positive sample weights increases with the number of epochs, which shows that the contribution of positive samples for training is increasing during training. Such observation verifies the necessity of sampling an unbalanced train batch and the effectiveness of our meta-miner.

\begin{figure*}[t]
  \centering
  \subfigure[KinFaceW-I]{
    \label{fig:roc:morph2}
    \includegraphics[width=0.45\linewidth]{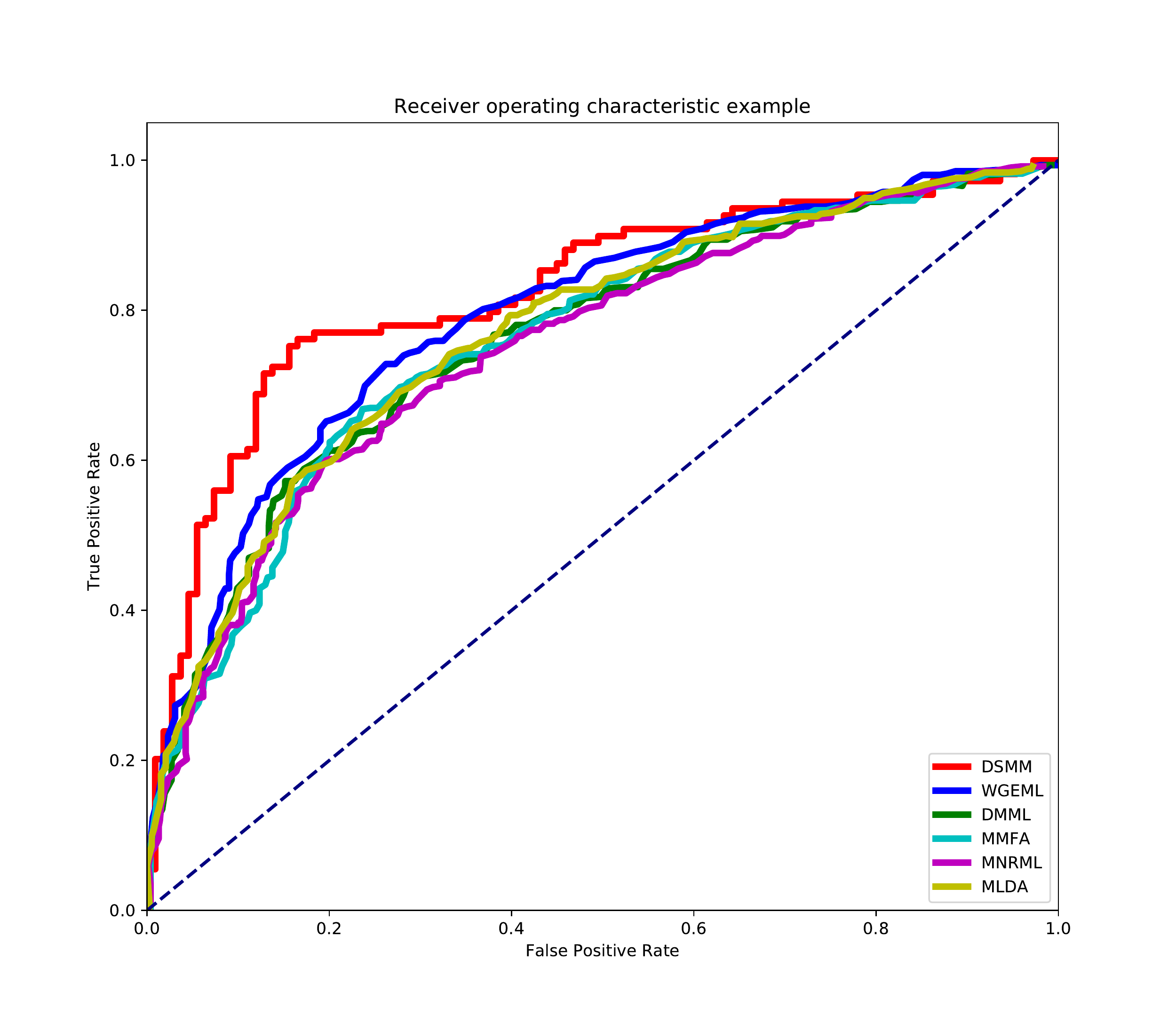}
  }
  \subfigure[KinFaceW-II]{
    \label{fig:roc:adience1}
    \includegraphics[width=0.45\linewidth]{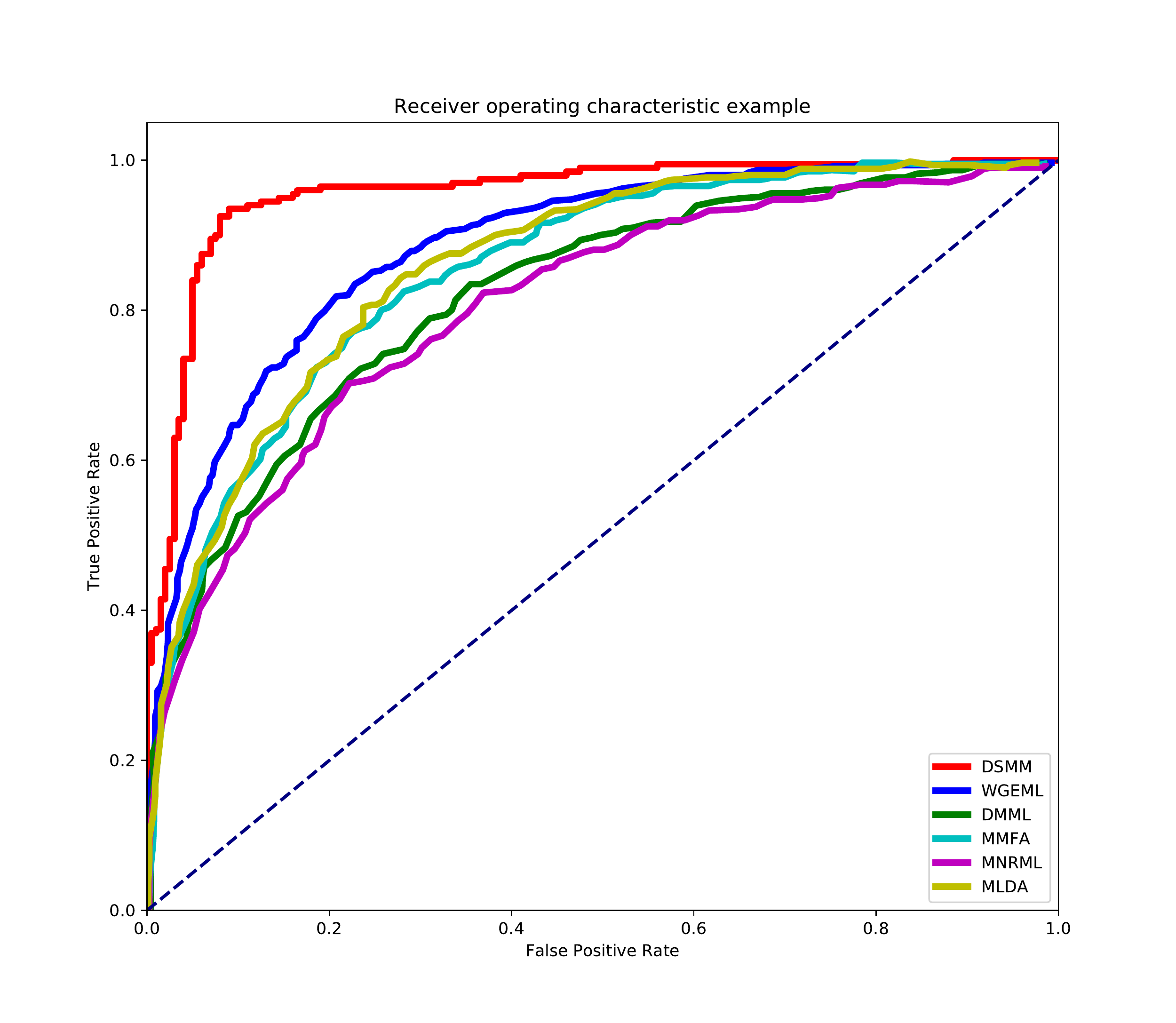}
  }
  \caption{The comparisons of ROC curves with other methods on the KinFaceW-I dataset and the KinFaceW-II dataset.}
  \label{fig:roc}
  \vspace{-0.5cm}
\end{figure*}

{\bf Parameters Analysis:}
We conducted experiments on the KinFaceW-I and KinFaceW-II datasets to choose the best value for parameter $C$. The results are shown in Table \ref{tab:KFW1C} and Table \ref{tab:KFW2C}, respectively. On the KinFaceW-I data set, we set the values of $C$ as 1, 2, 4, and 8, respectively. The experimental results show that our method achieves the best performance when $C=4$. On the KinFaceW-II data set, we set the values of $C$ as 1, 2, 3, and 4, respectively. The experimental results show that our method achieves the best performance when $C=2$. The above setting is used for our following experiments on the corresponding datasets.

%\begin{figure*}[t]
%  \centering
%  \subfigure[KinFaceW-I]{
%    \label{fig:corre:morph2}
%    \includegraphics[width=0.47\linewidth]{loss_curve_1.pdf}
%  }
%  \subfigure[KinFaceW-II]{
%    \label{fig:corre:adience1}
%    \includegraphics[width=0.47\linewidth]{loss_curve_2.pdf}
%  }
%  \caption{The loss curves on the KinFaceW-I dataset and KinFaceW-II dataset.}
%  \label{fig:ex2}
%
%\end{figure*}

\begin{table}
\centering
\caption{Comparison of verification accuracy of DSMM with state-of-the-art on the
KinFaceW-I dataset.}
\label{tab:SOTAKFW1}
\vspace{0.2cm}
\begin{tabular}{cccccc}
\toprule
Method& FD & FS & MD & MS & Mean\\
\midrule
IML~\cite{yan2014discriminative}&67.5&70.5&72.0&65.5&68.9 \\
MNRML~\cite{lu2014neighborhood}&66.5&72.5&72.0&66.2&69.3 \\
MPDFL~\cite{yan2014prototype}&73.5&67.5&66.1&73.1&70.1 \\
DMML~\cite{yan2014discriminative}&69.5&74.5&75.5&69.5&72.3 \\
GA~\cite{dehghan2014look}&76.4&72.5&71.9&77.3&74.5 \\
WGEML~\cite{liang2018weighted}&{\bf78.5}&73.9&80.6&81.9&78.7 \\
WLDA~\cite{sharma2012generalized}&76.6&71.2&77.7&76.4&75.5 \\
MMFA~\cite{sharma2012generalized}&77.9&72.0&77.2&75.2&75.6 \\
CNNP~\cite{zhang122015kinship}&71.8&76.1&84.1&78.0&77.5 \\
\midrule
DSMM& 76.7 & {\bf81.7} & {\bf89.0} & {\bf82.3} & {\bf82.4} \\
\bottomrule
\end{tabular}
\end{table}

\begin{table}
\centering
\caption{Comparison of verification accuracy of DSMM with state-of-the-art on the
KinFaceW-II dataset.}
\label{tab:SOTAKFW2}
\vspace{0.2cm}
\begin{tabular}{cccccc}
\toprule
Method& FD & FS & MD & MS & Mean\\
\midrule
IML~\cite{yan2014discriminative}&74.0&74.5&78.5&76.5&75.9 \\
MNRML~\cite{lu2014neighborhood}&74.3&76.9&77.6&77.4&76.6 \\
MPDFL~\cite{yan2014prototype}&77.3&74.7&77.8&78.0&77.0 \\
DMML~\cite{yan2014discriminative}&76.5&78.5&79.5&78.5&78.3 \\
LM$^{3}$L~\cite{hu2014large}&82.4&74.2&79.6&78.7&78.7 \\
GA(~\cite{dehghan2014look})&83.9&76.7&83.4&84.8&82.2 \\
WGEML~\cite{liang2018weighted}&88.6&77.4&83.4&81.6&82.8 \\
WLDA~\cite{sharma2012generalized}&86.6&74.4&81.0&78.8&80.2 \\
MMFA~\cite{sharma2012generalized}&85.6&73.2&80.4&77.2&79.1 \\
CNNP~\cite{zhang122015kinship}&81.9&89.4&92.4&89.9&88.4 \\
\midrule
DSMM& {\bf89.8} & {\bf92.6} & {\bf95.8} & {\bf93.6} & {\bf93.0} \\
\bottomrule
\end{tabular}
\vspace{-0.5cm}
\end{table}

\begin{table}
\centering
\caption{Comparison of verification accuracy of DSMM with state-of-the-art on the
TSKinFace dataset.}
\label{tab:SOTATSK}
\vspace{0.2cm}
\begin{tabular}{cccccc}
\toprule
Method& FD & FS & MD & MS & Mean\\
\midrule
BSIF~\cite{wu2016usefulness} & 81.4 & 81.5 & 79.9 & 82.0 & 81.2 \\
DDMML~\cite{lu2017discriminative} & 86.6 & 82.5 & 83.2 & 84.3 & 84.2 \\
GMP~\cite{zhang2015group} & 88.5 & 87.0 & 87.9 & 87.8 & 87.8 \\
MKSM~\cite{zhao2018learning} & 82.0 & 81.4 & 82.3 & 81.9 & 81.9 \\
\midrule
DSMM & {\bf91.4} & {\bf92.4} & {\bf93.9} & {\bf93.2} & {\bf92.7} \\
\bottomrule
\end{tabular}
\end{table}

\begin{table}
\centering
\caption{Comparison of verification accuracy of DSMM with state-of-the-art on the
Cornell dataset.}
\label{tab:SOTACor}
\vspace{0.2cm}
\begin{tabular}{cccccc}
\toprule
Method& FD & FS & MD & MS & Mean\\
\midrule
PSM~\cite{fang2010towards} & 72.9 & 54.6 & 73.8 & 61.3 & 70.7 \\
MNRML~\cite{lu2013neighborhood} & 74.5 & 68.8 & 77.2 & 65.8 & 71.6 \\
MPDFL~\cite{yan2014prototype} & 74.8 & 69.1 & 77.5 & 66.1 & 71.9 \\
DMML~\cite{yan2014discriminative} & 76.0 & 70.5 & 77.5 & 71.0 & 73.5 \\
MKSM~\cite{zhao2018learning} & 80.5 & 80.6 & 79.5 & 86.2 & 81.7 \\
\midrule
DSMM & {\bf88.0} & {\bf87.0} & {\bf94.0} & {\bf93.0} & {\bf90.5} \\
\bottomrule
\end{tabular}
\vspace{-0.5cm}
\end{table}

{\bf Comparisons with State-of-the-Art:}
We compared the proposed DSMM with existing kinship verification methods on the KinFaceW-I database, the KinFaceW-II database, the TSKinFace database, and the Cornell Kinship datasets.
%Some competitive methods include MNRML~\cite{lu2013neighborhood}, MPDFL~\cite{yan2014prototype}, DMML~\cite{yan2014discriminative}, MKSM~\cite{zhao2018learning}, BSIF~\cite{wu2016usefulness}, DDMML~\cite{lu2017discriminative}, GMP~\cite{zhang2015group}, GA~\cite{dehghan2014look}, CNNP~\cite{zhang122015kinship}, and LM$^{3}$L~\cite{hu2014large}.

Table~\ref{tab:SOTAKFW1} and Table~\ref{tab:SOTAKFW2} show the experimental results on the KinFaceW-I and KinFaceW-II databases respectively. Bolded numbers represent the best results. On the KinFaceW-I data set, the average verification rate of our proposed method is 82.4\%, which outperforms the previous best method  WGEML~\cite{liang2018weighted} by 3.7\%. The proposed method achieves an average verification rate of 93.0\% on the KinFaceW-II data set, which outperforms the previous best approach  CNNP~\cite{zhang122015kinship} by 4.6\%.

We also present the comparisons of our DSMM method and other methods on the KinFaceW-I and KinFaceW-II datasets in terms of ROC curves. As can be seen in Figure \ref{fig:roc}, the curve of the DSMM method is higher than other methods, which demonstrates the superiority of the proposed discriminative sample meta-mining approach. Note that the KVRL~\cite{kohli2016hierarchical} achieves higher scores due to the multi-patches input and outside data training. For a fair comparison, we do not include it in Table~\ref{tab:SOTAKFW1} and Table~\ref{tab:SOTAKFW2}.

Table \ref{tab:SOTATSK} shows the experimental results on the TSKinFace database while Table \ref{tab:SOTACor} lists the experimental results on the Cornell  KinFace database. We observe that our method attains an average verification rate of  92.7\% on the TSKinFace dataset, outperforming the GMP~\cite{zhang2015group} method by 4.9\%. The proposed discriminative sample meta-mining method obtains an average verification rate of 90.5\% on the Cornell dataset, which outperforms the previous best-performing  MKSM~\cite{zhao2018learning} method by 8.8\%.

%
%\textbf{Convergence Analysis:}
%To further show the convergence of our method, we show the loss curves on KinFaceW-I and KinFaceW-II database in Figure \ref{fig:ex2}. We observe that our method converges  well on both databases.

\section{Conclusion}

In this paper, we have presented the discriminative sample meta-mining approach for kinship verification. Unlike the existing methods that usually construct a balanced database with fixed negative samples, our method utilizes all possible pairs and aims to exploit the discriminative information from the limited positive samples and sufficient negative pairs. For each iteration, an unbalanced train batch and a balanced meta-train batch are sampled. The balanced meta-train batch is used to guide the training process of our meta-miner via a meta-learning framework. Then our kinship model is trained with the discriminative samples on the training batch, which are weighted by the meta-miner. Extensive experiment results on four widely used kinship databases demonstrate that our method significantly outperforms the state-of-the-art methods.

\section*{Acknowledgement}
This work was supported in part by the National Natural Science Foundation of China under Grant 61822603, Grant U1813218, Grant U1713214, in part by Beijing Academy of Artificial Intelligence (BAAI), and in part by a grant from the Institute for Guo Qiang, Tsinghua University.

{\small
\bibliographystyle{ieee_fullname}
\bibliography{egbib}
}

\end{document}